\documentclass[letterpaper, 10 pt, conference]{ieeeconf}  

\IEEEoverridecommandlockouts                              

\usepackage{graphics} 
\usepackage{epsfig} 
\usepackage{mathptmx} 
\usepackage{times} 
\usepackage{amsmath} 
\usepackage{amssymb}  
\usepackage{textcomp}
\usepackage{gensymb} 
\usepackage{float}
\usepackage{siunitx}
\usepackage{comment}
\usepackage{bm}
\usepackage{booktabs}

\title{\LARGE \bf
 Hydrodynamic Performance Enhancement of Unmanned Underwater Gliders with Soft Robotic Morphing Wings
 for Agility Improvement
}
\author{A. Giordano$^{1,5}$, G. De Meurichy$^{1,*}$, V. Telazzi$^{1}$, C. Mucignat$^{3}$, I. Lunati$^{3}$, D. A. L. M.  Louchard$^{1}$, M. Iovieno$^{4}$,\\ S. F. Armanini$^{5}$, and M. Kovac$^{1,2,*}$
\thanks{$^{1}$ A. Giordano, G. De Meurichy, V. Telazzi, D. A. L. M.  Louchard, and M. Kovac are with the Laboratory of Sustainability Robotics, Empa, 8600 Dübendorf, Switzerland. {\tt\small \{gregory.demeurichy\}@gmail.com}}
\thanks{$^{2}$ M. Kovac is with
EPFL, 1005 Lausanne, Switzerland. {\tt\small \{mirko.kovac\}@epfl.ch}}
\thanks{$^{3}$ C. Mucignat, and I. Lunati are with the Laboratory for Computational Engineering, Empa, 8600 Dübendorf, Switzerland.}
\thanks{$^{4}$ M. Iovieno is with the Dipartimento di Ingegneria Meccanica e Aerospaziale (DIMEAS), Politecnico di Torino, 10129 Torino, Italy.} 
\thanks{$^{5}$ A. Giordano, and S. F. Armanini are with the Department of Aeronautics, Imperial College London, SW7 2AZ London, United Kingdom.} 
\thanks{$^{*}$ Correspondence should be addressed to these authors.}}
\begin{document}

\maketitle
\thispagestyle{empty}
\pagestyle{empty}
\begin{abstract}
This work aims to assess the hydrodynamic efficiency of Underwater Unmanned Vehicles (UUVs) equipped with soft morphing wings in comparison with conventional rigid ones. Unlike rigid wings, deformable counterparts are capable of altering their aerodynamic properties on demand. Any enhancement in a UUV’s hydrodynamic efficiency extends its operational range, which may be a determining factor for mission feasibility. Structural and Computational Fluid Dynamics (CFD) simulations have been conducted for both a soft morphing wing and a UUV incorporating it. The results are validated against experimental data available in the literature. The simulated soft wing reproduces the experimental deformation with an average error of 0.752 mm, corresponding to 0.327\% of its chord length, and the lift-to-drag ratio measured in water tunnel experiments with an average error of 15.7\%. Finally, the overall efficiency of a UUV employing soft wings is found to be 9.75\% higher than that of an equivalent UUV with traditional rigid wings. These findings further confirm the potential of soft robotics to enhance the performance of underwater vehicles, particularly in applications where pressure agnosticity is required.
\end{abstract}
\section{Introduction}
Unmanned Underwater Vehicles (UUVs) are submarine gliders primarily employed for the environmental monitoring of oceans~\cite{javaid2017effect}. They require no active propulsion, but vary their buoyancy to induce vertical motion and, through their wings, convert it into a sinusoidal trajectory. Where the ocean floor permits, the amplitude of a UUV’s path may reach several thousand metres, its wavelength a few kilometres, with an overall mission range of up to $1000$~km~\cite{fan2014dynamics, singh2017cfd, suberg2014assessing}. Consequently, the wings are major contributors to the total hydrodynamic efficiency of UUVs. Since energy storage remains a bottleneck for untethered robotic systems, wing performance ultimately determines whether long-range missions can be executed at all. Indeed, for exploration in cluttered environments such as the Arctic or the Southern Ocean, where the deployment or recovery of a UUV may be hindered by pack ice or icebergs, operational range is a critical factor~\cite{lucas2025giant}.
Conventional UUVs typically employ rigid fixed wings with a symmetric hydrofoil, a small angle of incidence, and no control surfaces. Due to the extreme pressure of the depths UUVs reach, dynamic seals are considered a vulnerability which, in case of failure, could compromise the entire vehicle and lead to its loss~\cite{wu2023numerical,wu2024effect}. On the other hand, soft actuators exhibit high resistance to extreme-pressure environments, making them effectively pressure-agnostic~\cite{Katzschmann2016HydraulicSwimming}. This paper investigates the implementation of soft morphing wings on UUVs, focusing on the hydrodynamic efficiency of the entire vehicle and hence also on its maximum achievable range. 
The proposed methodology comprises the following steps:
\begin{enumerate}
    \item simulating the structural and hydrodynamic behaviour of a soft morphing wing previously tested in the literature~\cite{Giordano2024}, and validating the simulations against the corresponding experimental data;
    \item simulating the hydrodynamic behaviour of a conventional UUV and validating the results against empirical data reported in the literature;
    \item simulating the hydrodynamic behaviour of a UUV incorporating the aforementioned soft wing and comparing its overall efficiency, defined as the lift-to-drag ratio, with that of an identical UUV equipped with a rigid wing.
\end{enumerate}
\begin{figure}[!tb]
    \centering
            \centering
                            
            \hskip -0.18in
            \includegraphics[width=3.54in]{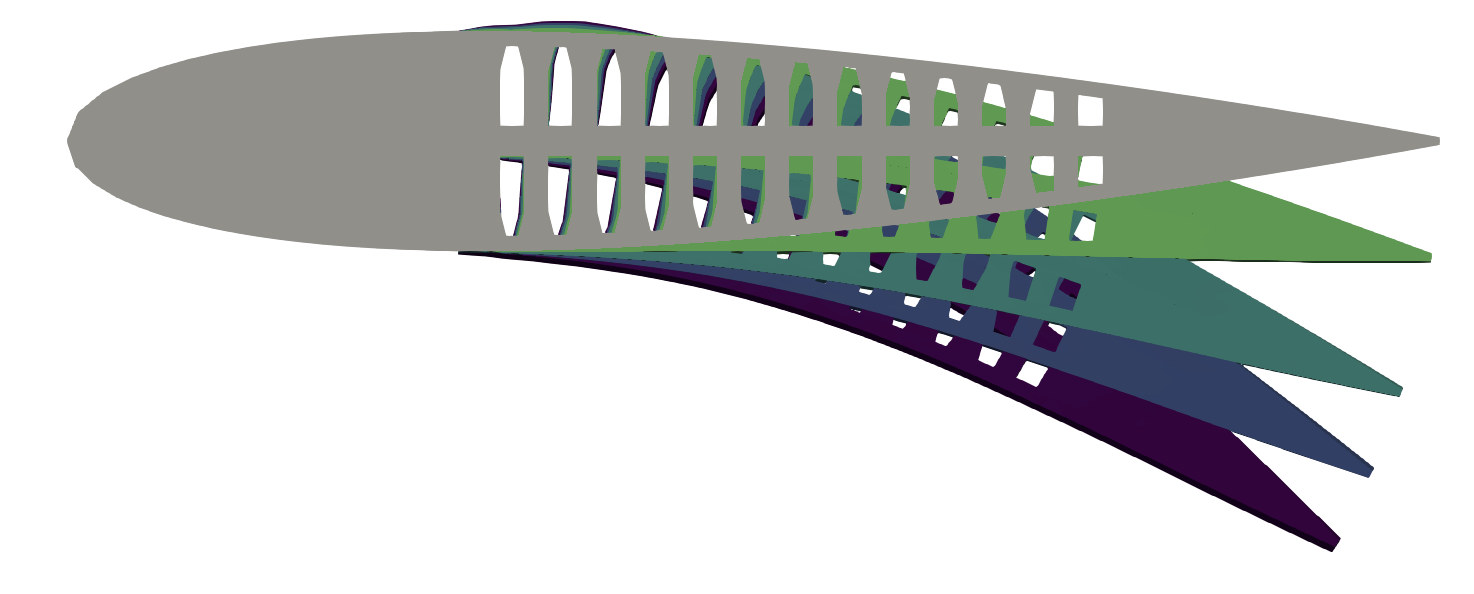}
            \vskip -2eX
    \caption{Cross-sectional view of the soft morphing wing showing its profile and internal structure for all deformation levels tested in~\cite{Giordano2024}. From top to bottom: \SI{0}{\milli\litre}, \SI{30}{\milli\litre}, \SI{60}{\milli\litre}, \SI{90}{\milli\litre}, and \SI{120}{\milli\litre}. All profiles shown are the results of the simulations performed for the present work.}
    \vskip -4eX
    \label{soft_wing_def}
\end{figure}

\section{Structural Simulation and Validation}
The soft robotic morphing wing presented in~\cite{Giordano2024} is selected for the purposes of the present study. The wing features a symmetric NACA0016 profile and two symmetric actuation chambers, located on the suction and pressure sides, and extending along the entire wingspan. The leading edge is manufactured from aluminium, while the remaining part is composed of EcoFlex 00-50 silicone (Smooth-On Inc., Macungie, USA). At rest, the chambers of the wing are filled with equal volumes of water. To generate an upward camber, a pump transfers water from the upper chamber to the lower one. The chambers are separated by an inextensible Poly Lactic Acid (PLA) layer, \SI{2}{\mm} thick, which enhances the structural stability of the wing. 

\begin{figure}[!t]
    \centering
            \centering
            \includegraphics[width=\columnwidth]{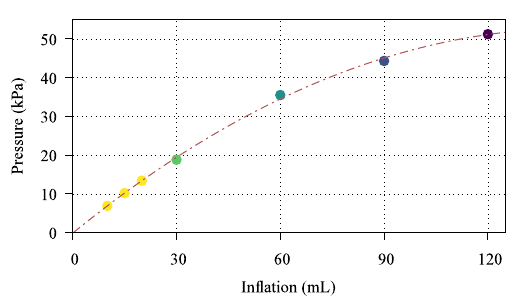}
                \vskip -2eX
\caption{Correlation between the inflation levels of the soft wing~\cite{Giordano2024} and the isotropic pressure field applied in the simulations to reproduce equivalent deformations. A second-order polynomial fit $P(I) = -0.025I^2 + 0.7273I - 0.1543$ was applied to the four experimental data points, enabling the extrapolation of 3 additional inflation levels (yellow).}
    \vskip -3eX
  \label{Pressure}
\end{figure}

In \cite{Giordano2024}, the wing has been submerged in a water channel to observe its structural deformation, and its hydrodynamic properties were measured using a load cell. The deformation was evaluated in terms of \textit{differential inflation}, i.e. the volume of water pumped from one chamber to the other (30, 60, 90, and \SI{120}{\milli\litre}). Fig.~\ref{soft_wing_def} illustrates the deformation of the soft wing corresponding to the inflations investigated in \cite{Giordano2024}. 
The hyperelastic behaviour of EcoFlex~00-50 was modelled using the two-parameter Mooney–Rivlin formulation:
$$
W = C_1 (I_1 - 3) + C_2 (I_2 - 3),
\eqno{(1)}
\label{eq:moon}
$$
where $W$ is the strain energy density of the material, $C_1$ and $C_2$ are empirically determined material constants~\cite{lavazza2023strain}, and $I_1$ and $I_2$ are the invariants of the left Cauchy–Green deformation tensor.
A complete list of the material parameters used in the simulation is provided in Table~\ref{tab:aluminium_ecoflex_pla}. 
Owing to its complex geometry, the wing was meshed using the \textit{Sweep method}, with the body sizing set to \SI{1e-2}{\metre}. In the initial condition, the soft wing is undeformed and at rest, whereas the boundary conditions are defined as follows:
\begin{enumerate}
    \item the aluminium leading edge is fixed in space, with bonded interfaces to the silicone wing and the inextensible layer, preventing interpenetration and relative motion;
    \item likewise, the interface between the silicone wing and the inextensible layer is bonded.
\end{enumerate}
The forcing term consists of an isotropic pressure field applied to one of the chambers, which is linearly increased from \SI{0}{\pascal} to its full value over the course of one second. The full pressure value is initially unknown, since in the previous experiments~\cite{Giordano2024} the extent of deformation was quantified in terms of differential inflation rather than the internal chamber pressure. Consequently, the input pressure was iteratively adjusted to achieve the closest wing camber to that reported in~\cite{Giordano2024}. Table~\ref{inflation_error} relates the differential inflation to the corresponding pressure applied in the simulations. A least-squares regression was performed on the pressure values, as shown in Fig.~\ref{Pressure}, and the resulting parabolic fit provides a reasonable interpolation owing to the strong non-linearity associated with hyperelastic materials. 
The deformation obtained from the simulations for each inflation level was compared with the experimental data presented in~\cite{Giordano2024} (Fig.~5) by superimposing the edge of the suction side at the lower surface of each wing. The comparison is illustrated in Fig.~\ref{inflations}, and the Root Mean Square (RMS) and maximum errors are reported in Table~\ref{inflation_error}. The average of all RMS errors is \SI{7.5e-3}{\metre}, which is negligible (\num{0.327}\%) when compared with the wing’s chord length, $c = \SI{0.230}{\metre}$.

\section{Wing CFD Simulation and Validation}
The deformed wing obtained from the structural simulations was imported into \textsc{Ansys Fluent} for Computational Fluid Dynamics (CFD) analyses. The simulations were conducted at free-stream velocities of $U_1 = \SI[per-mode=symbol]{0.25}{\metre\per\second}$ and $U_2 = \SI[per-mode=symbol]{0.40}{\metre\per\second}$, as they were tested experimentally in~\cite{Giordano2024}. 

\begin{figure}[!bt]
    \centering
            \centering
            \includegraphics[width=\columnwidth]{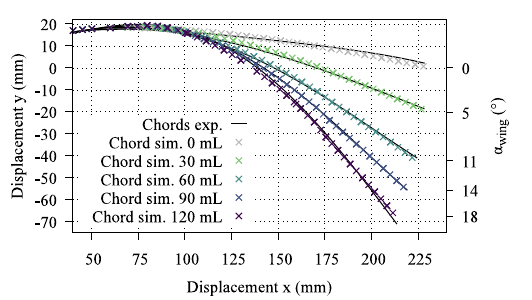}
             \vskip -2eX
    \caption{Camber profiles of the soft wing from the experiments performed in~\cite{Giordano2024}, Fig.~5 (solid black lines), and corresponding profiles as an output of the structural simulations performed for the present work (coloured dots).} 
    \vskip -4eX
    \label{inflations}
\end{figure}

In addition, the cruising velocities of UUVs reported in the literature generally fall within this range. 
A separate simulation was performed for each combination of 
angle of attack (AoA) $\alpha$ and inflation level tested in the water tunnel experiments of~\cite{Giordano2024}, resulting in a total of 35 simulations per flow velocity. Following~\cite{Giordano2024}, in the present analysis AoA is defined with respect to the wing's original symmetric profile, and is therefore kept constant during camber morphing. However, wing inflation modifies the camber and, according to the classical aerodynamic definition of AoA, effectively rotates the chord line, leading to a change in the actual AoA experienced by the wing. Fig.~\ref{inflations} illustrates the correlation between camber morphing and the AoA in its classical aerodynamic definition (denoted as $\alpha_{wing}$). To characterise the hydrodynamic regime at these flow conditions, an evaluation of the Reynolds number was carried out. The Reynolds number, $Re$, is a dimensionless similitude parameter defined as the ratio between inertial and viscous forces:
$$
\text{Re} = \frac{U L}{\nu}
\eqno{(2)}
\label{eq:re}
$$
where $\nu = \SI[per-mode=symbol]{1.00e-6}{\metre\squared\per\second}$ is the kinematic viscosity of water, $U$ the free-stream velocity, and $L$ the characteristic length of the problem (in this case, the wing’s chord length). Consequently, $U_1$ corresponds to $\text{Re}_1 \simeq \num{5.75e4}$, and $U_2$ to $\text{Re}_2 \simeq \num{9.20e4}$. Most of the domain is discretised using tetrahedral elements with an average size of \SI{7e-2}{\metre}, whereas the region in the vicinity of the wing is refined with tetrahedral elements of average dimension \SI{6e-3}{\metre}. To resolve the boundary layer, a ten-layer inflation structure was adopted, with an initial layer height of \num{8e-5}~m.
\begin{table}[!b]
\vskip -3eX
\caption{Material Properties of Aluminium, Ecoflex 00-50, and PLA}
\vskip -3eX
\label{tab:aluminium_ecoflex_pla}
\begin{center}
\setlength{\tabcolsep}{5pt}
\begin{tabular}{lS[table-format=-1.1e-1]
S[table-format=-1.1e-1]
S[table-format=-1.1e-1]}
\toprule
\textbf{Property} & \textbf{Aluminium} & \textbf{Ecoflex 00-50} & \textbf{PLA} \\
\midrule
Density [kg m$^{-3}$] & 2.70e3 & \num{1.07e3} & \num{1250} \\
Young's Mod. [MPa] & 6.90E4 & \multicolumn{1}{c}{--} & 3.50e3 \\
Poisson's Ratio & 3.30e-1 & \multicolumn{1}{c}{--} & 2.00e-1 \\
Bulk Mod. [MPa] & 6.96e4 & \multicolumn{1}{c}{--} & 1.94e3 \\
Shear Mod. [MPa] & 2.60e4 & \multicolumn{1}{c}{--} & 1.46e3 \\
$C_{10}$ [MPa] & \multicolumn{1}{c}{--} & 4.76e-2 & \multicolumn{1}{c}{--} \\
$C_{01}$ [MPa] & \multicolumn{1}{c}{--} & 1.19e-2 & \multicolumn{1}{c}{--} \\
$I_1 = I_2$& \multicolumn{1}{c}{--} & 1.00e-2 & \multicolumn{1}{c}{--} \\
\bottomrule
\end{tabular}
\end{center}
\vskip -3eX
\end{table}

\begin{table}[h] 
    \caption{Comparison between inflation, input pressure, root mean square (RMS) and maximum (max) error.}
    \vskip -3eX
    \label{inflation_error}
    \begin{center} 
    \setlength{\tabcolsep}{7pt}
    \begin{tabular}{cccc} 
    \toprule 
    \textbf{Inflation } & \textbf{RMS Error} & \textbf{Max Error} & \textbf{Pressure} \\ 
    $\text{[}$mL] & [mm] & [mm] & [kPa] \\
    \midrule 
     ~~0  & 0.921 & 1.60~$\,$ & 0.00 \\ 
    ~30  & 0.835 & 1.26~$\,$ & 18.8 \\ 
    ~60  & 0.632 & 1.05~$\,$ & 35.5 \\ 
    ~90  & 0.301 & 0.543 & 44.3 \\ 
    120 & 1.07~$\,$ & 1.60~$\,$ & 51.2 \\ 
    \bottomrule 
    \end{tabular}
    \end{center} 
        \vskip -6eX
\end{table}
%
%
The initial conditions comprise a free-stream velocity of $U_1$ or $U_2$ at the inlet and an isotropic pressure equal to \SI{1e5}{\pascal}. 
For both Reynolds numbers, the Reynolds-Averaged Navier–Stokes (RANS) equations were solved using the $k$–$\omega$ Shear-Stress Transport (SST) turbulence model to ensure proper resolution of transient flow features, laminar flow separation, and subsequent turbulent reattachment. The model converged after approximately 1500 iterations.
At $U_1$, the flow around the wing transitions from laminar to turbulent and is characterised by the formation of a laminar separation bubble (LSB), which is known to induce strong sensitivity to experimental conditions and turbulence modelling choices \cite{jardin2025naca0012}. As a result, for $\text{Re}_1 \simeq \num{5.75e4}$, the computed lift and drag deviate from the experimental data in \cite{Giordano2024} by 50–90\%. By contrast, at $\text{Re}_2 \simeq \num{9.20e4}$ the flow is further from the unstable transitional regime, and the predicted lift and drag show significantly improved agreement with experiments (average error of 15.7\%).
Interestingly, a symmetric wing profile immersed in a transitional flow at $\alpha = 0\degree$ can generate lift as a result of the laminar separation bubble, and a symmetric wing with a small but positive AoA may exhibit negative lift~\cite{aguilar2020onset}. This feature, together with the systematic errors reported by the authors, explains why the $C_L$ curve of the uncambered symmetric wing presented in~\cite{Giordano2024} does not intersect the origin.
While absolute force predictions in transitional flows remain model-dependent, the relative performance trends between rigid and morphing configurations were found to be consistent across all tested conditions and therefore provide confidence in the methodology employed and form the basis of the conclusions of this work.
\section{Glider CFD Simulations and Validation}
CFD simulations to characterise the hydrodynamic performance of a UUV,  validated against corresponding experimental results, are reported in~\cite{singh2017cfd} and~\cite{ichihashi2008development}. In the present study, we validate our CFD methodology by reproducing an identical UUV in \textsc{Ansys Fluent}. The same simulations are then repeated, this time implementing the soft morphing wing on the UUV.
\subsection{CFD Validation}
To validate our CFD methodology against the simulations and experiments presented in \cite{singh2017cfd} and \cite{ichihashi2008development}, we implemented an identical UUV (Fig. \ref{fig:uuv}), including its NACA0009 wing at the same $Re$. For validation, two free-stream flow speeds are chosen: $U_1 = 0.50$ \unit{\metre\per\second}, as in \cite{singh2017cfd}, and \cite{ichihashi2008development}, and $U_2 = 0.26$ $m/s$, as in \cite{Giordano2024}. When using $U_2$, a proportional scale-up of the UUV was required to match the $Re$ in \cite{singh2017cfd} and \cite{ichihashi2008development}, as shown in 3 and 4. 
$$
\text{Re} = \frac{U_1\, L_1}{\nu} = \frac{0.50 \text{ $m/s$ } \cdot 0.213 \unit{\metre}}{\num{1.00e-6} \unit{\metre\squared\per\second}} \simeq \num{1.06e5} \eqno{(3)} 
\label{eq:re2}
$$

$$
L_2 = \frac{\text{Re}\,\nu}{U_2} = \frac{\num{1.06e5} \cdot \num{1.00e-6} \unit{\metre\squared\per\second}}{\SI{0.26}{\metre\per\second}} \simeq \SI{0.409}{\metre}
\eqno{(4)}
\label{eq:re3}
$$
To ensure an adequate representation of the flow phenomena in each simulation involving the above-mentioned UUV, the computational domain shown in Fig.~\ref{fig:uuv_domain} was employed. A structured mesh of tetrahedral elements was generated, with a maximum element size of \SI{9.5e-2}{\metre}.

\begin{figure*}[!tb]
    \centering
    \includegraphics[width=\textwidth]{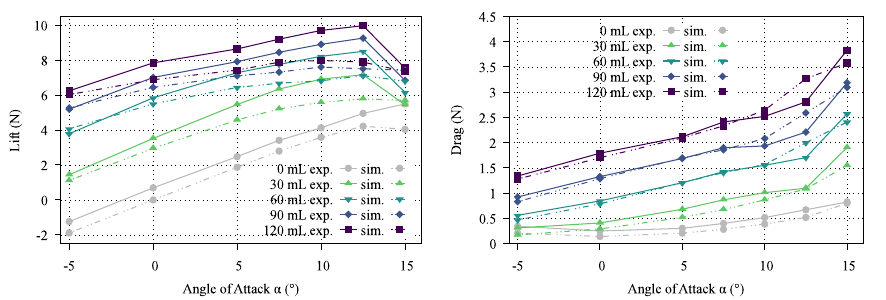}
    \caption{Hydrodynamic characterisation of the soft wing at $Re_2 \simeq \num{9.20e4}$ (free-stream velocity $U_2 = \SI[per-mode=symbol]{0.40}{\metre\per\second}$). Solid lines represent the water tunnel measurements, as reported in~\cite{Giordano2024}, while the dotted lines correspond to the force values obtained from the CFD simulations performed for the present work.}
    \label{fig:liftdragwide}
    \vskip -3eX
\end{figure*}

\begin{figure}[!tb]
    \centering
            \includegraphics[width=\columnwidth]{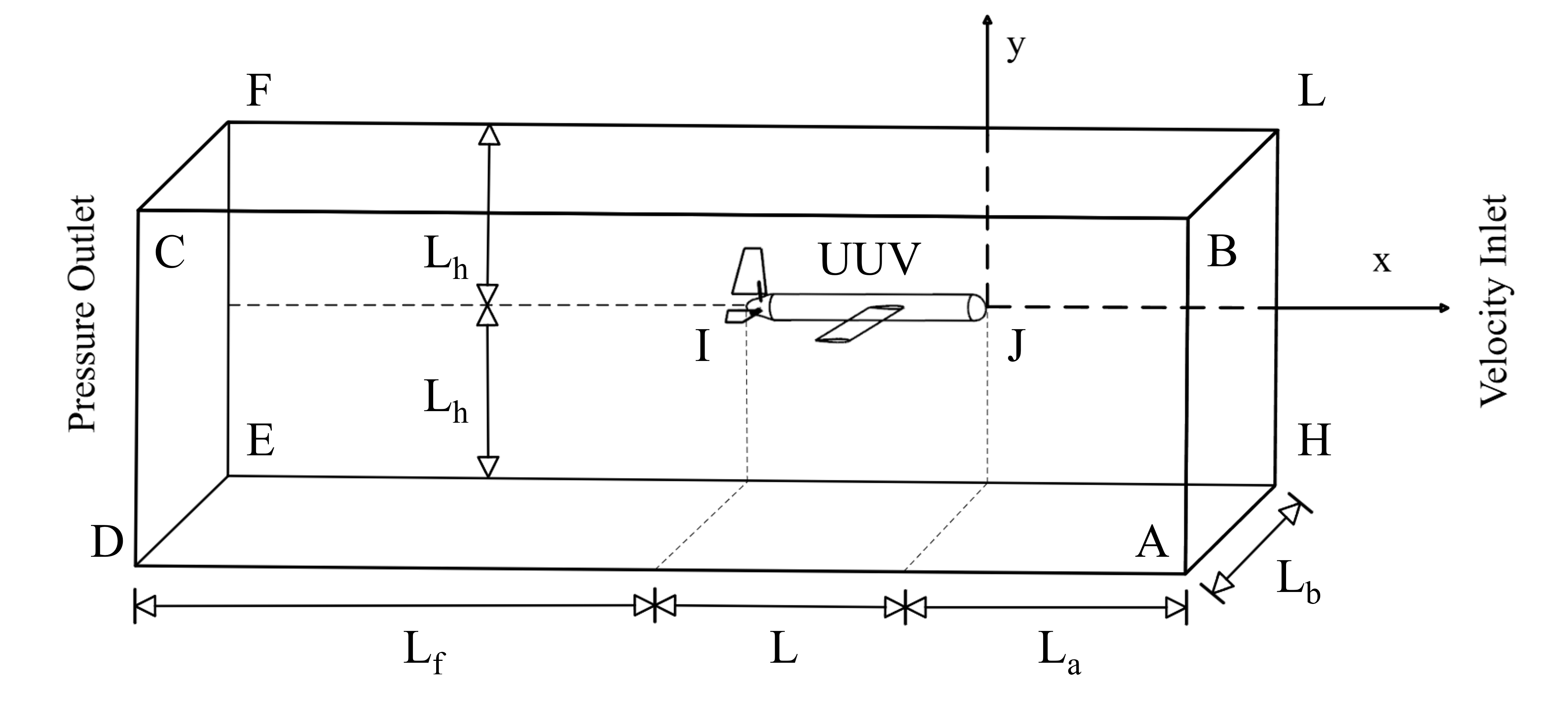}
            \vskip -2eX
    \caption{Domain of all CFD simulation performed by the authors and involving an underwater glider. The cylinder represents the UUV and has a total length of $\overline{IJ} = L_{UUV}$. The parallelepiped of vertices A, B, C, D, E, F, G, and H delimits the water volume. $L_a = L_b = L_h = 1.5\cdot L_{UUV}$, $L_f = 5 \cdot L_{UUV}$.}
    \vskip -1eX
    \label{fig:uuv_domain}
\end{figure}
At $\text{Re} = \num{1e5}$, the flow is fully turbulent. Accordingly, the RANS equations with the $k$–$\omega$~SST turbulence model were adopted, given its widespread use and proven reliability in similar applications. The initial conditions consisted of a free-stream velocity $U_i$ at the inlet and an AoA $\alpha$. The boundary conditions were specified as \textit{no-slip} on the vehicle’s surface and \textit{slip} at the outer boundaries of the domain. 
The results are presented in Fig.~\ref{fig:uuvliftdrdag}, which compares the coefficients of lift ($C_L$) and drag ($C_D$) obtained in the present simulations with the numerical and experimental studies reported in~\cite{singh2017cfd} and~\cite{ichihashi2008development}. Those coefficients were calculated from their definitions, 
$$
C_L = \frac{|\bm{L}|}{\tfrac{1}{2}\rho {U}^2 S}, \qquad C_D = \frac{|\bm{D}|}{\tfrac{1}{2}\rho {U}^2 S}. \eqno{(5)}
$$
\noindent
The coefficients $C_L$ and $C_D$ are plotted as functions of the AoA, ranging from \ang{-8} to \ang{+8}. For both $U_1$ and $U_2$, the lift results are in good agreement with the simulation and experimental data reported in~\cite{singh2017cfd} and~\cite{ichihashi2008development}, as the lift in Fig.~\ref{fig:uuvliftdrdag} exhibits the same linear trend. The same general behaviour is observed for $C_D$, although the agreement is slightly poorer: the simulations tend to overestimate the drag coefficient by an average absolute error of \num{0.015} with respect to the mean experimental value $\overline{C_D} \simeq \num{0.04}$.
\begin{figure}[t]
    \centering
            \centering
            \includegraphics[width=3in]{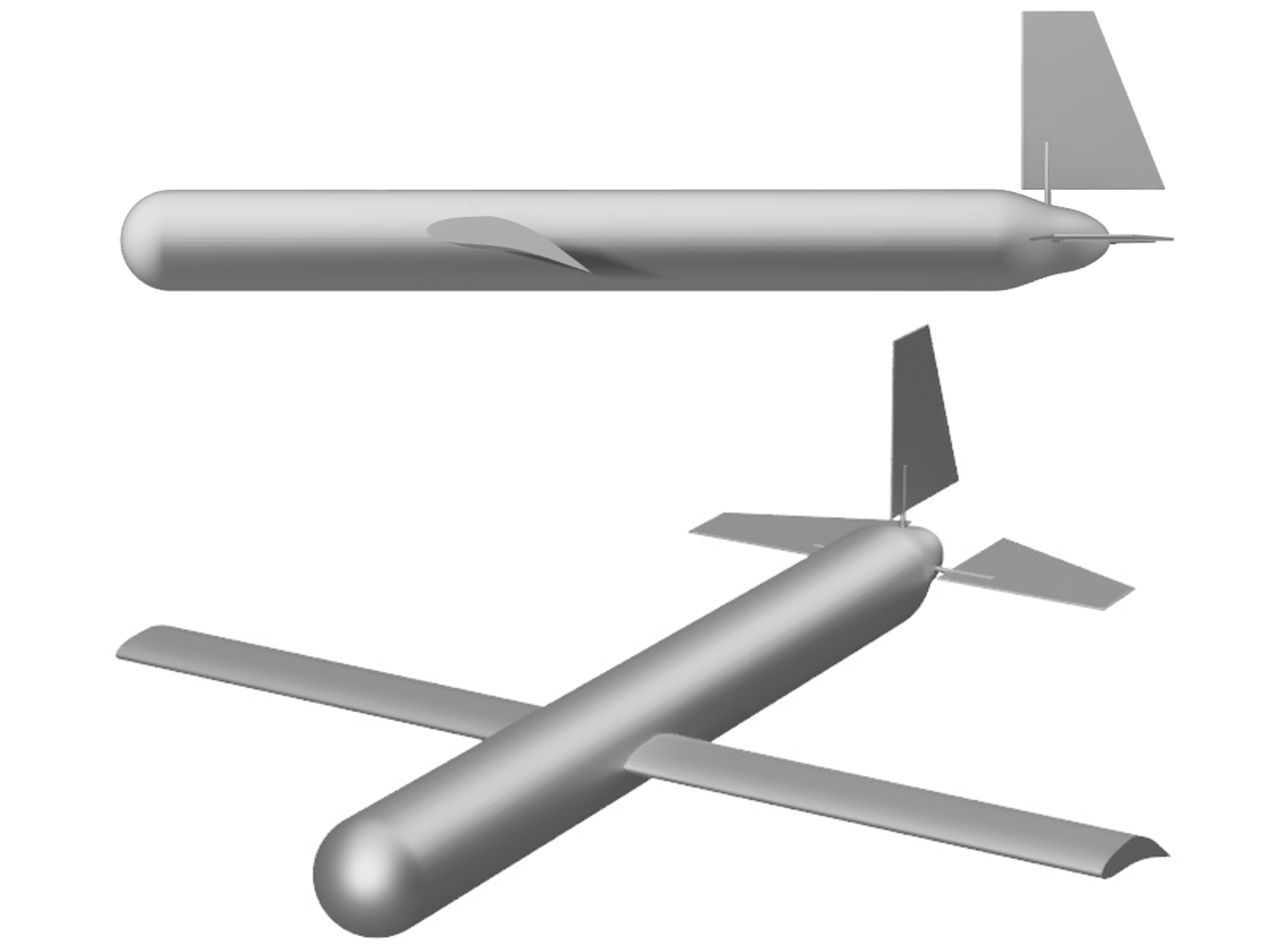}
    \caption{Two views of the 3D rendering of the simulated Underwater Unmanned Vehicle (UUV) implementing the soft wing.}
    \vskip -4eX
    \label{fig:uuv}
\end{figure}

\subsection{Soft wing implementation}
After validating the methodology, additional simulations were carried out on a UUV equipped with the soft morphing wing. For consistency and reproducibility, the same Reynolds number, computational domain, mesh settings, initial and boundary conditions, and solver configuration as those used in the validation simulations were employed. The same UUV model was retained, with only its original rigid wing replaced by the soft morphing one. While the airfoil profile remains identical to that reported in~\cite{Giordano2024}, the wingspan was extended to  \SI{0.714}{\metre} to match that of the scaled-up UUV used for validation. 
The incidence angle, defined as the angle between the chord line of the undeformed wing and the longitudinal axis of the fuselage, was fixed at \ang{0}. CFD simulations were then performed for the UUV featuring the soft wing across all deformation levels tested in~\cite{Giordano2024}, together with an additional interpolated deformation of \SI{15}{\milli\litre} (Fig.~\ref{Pressure}). Four operational flow speeds were examined: 0.15, 0.26, 0.35, and \SI[per-mode=symbol]{0.55}{\metre\per\second}, for AoA ranging from \ang{-8} to \ang{+8}. 
The flow speed of \SI[per-mode=symbol]{0.26}{\metre\per\second} was selected in the present study for validating the simulations against the results of~\cite{ichihashi2008development} and~\cite{singh2017cfd}. This velocity also closely corresponds to the \SI[per-mode=symbol]{0.25}{\metre\per\second} used in~\cite{Giordano2024}, for which extensive water-tunnel experiments were performed.
\begin{figure*}[!tb]
    \centering
    \includegraphics[width=\textwidth]{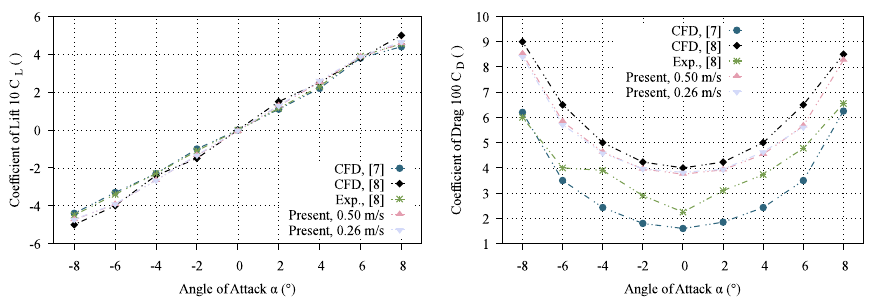}
    \caption{CFD validation of the simulations using the Underwater Unmanned Vehicle (UUV) presented in \cite{singh2017cfd} and \cite{ichihashi2008development}.}
    \vskip -3eX
    \label{fig:uuvliftdrdag}
\end{figure*}
Additionally, a flow speed of \SI[per-mode=symbol]{0.35}{\metre\per\second} corresponds to the operational velocity of UUVs reported in~\cite{hossain2025recent}. Simulations performed at \SI[per-mode=symbol]{0.15}{\metre\per\second} and \SI[per-mode=symbol]{0.55}{\metre\per\second} were also included to provide a comprehensive overview of the vehicle’s performance across a broader range of flow conditions. 
The following section presents the outcomes of these simulations through an analysis of the UUV’s hydrodynamic efficiency $\eta$, i.e. the ratio between its lift $\bm{L}$ and its drag $\bm{D}$ \cite{Bianco2003a}:
$$
\eta = \frac{|\bm{L}|}{|\bm{D}|} = \frac{C_L}{C_D}.
\eqno{(7)}
$$
For a flow speed of \SI[per-mode=symbol]{0.26}{\metre\per\second}, the hydrodynamic efficiency of the UUV is shown in Fig.~\ref{fig:efficiency026}. The UUV featuring the undeformed (hereafter referred to as \textit{rigid}) wing exhibits an efficiency curve that is symmetric with respect to the origin. In contrast, all other inflation levels produce asymmetric curves, owing to the geometric characteristics of the cambered profile. Achieving a symmetric efficiency in the domain of negative AoAs would require deformation in the opposite direction (defined as \textit{negative inflation} in \cite{Giordano2024}).
For higher AoAs, the \SI{15}{\milli\litre} inflation curve yields similar or superior efficiency, reaching a maximum $\eta = 6.98$ at $\alpha = \ang{6}$, \num{9.75}\% higher than the maximum $\eta = 6.36$ at $\alpha = \ang{8}$ for the rigid configuration. Conversely, at $\alpha = \ang{8}$, the \SI{15}{\milli\litre} curve decreases to an efficiency of 5.42. 
At higher inflation levels, the camber effect smooths the efficiency of the UUV across the entire analysed range. For the most inflated configurations (60, 90, and \SI{120}{\milli\litre}), the efficiency never falls below zero, indicating that the lift remains positive even at negative AoAs. A well-known consequence of camber is the shift of the linear trend of $C_L$ towards lower AoAs, accompanied by the earlier onset of stall~\cite{Bianco2003a}. For this reason, all inflated curves in Fig.~\ref{fig:efficiency026} exhibit a slight flattening at higher AoAs, whereas this behaviour is not yet evident for the rigid configuration. Furthermore, a similar trend was observed at all other tested flow speeds. 
Regarding static stability, the pitching moment of the UUV was computed for all inflations reported in~\cite{Giordano2024} and for AoAs between \ang{-8} and \ang{+8}. The pitch plot in Fig.~\ref{fig:piroll} assumes a uniform density distribution across the entire UUV. Under this assumption, the rigid configuration cannot satisfy the conditions for static stability, as its wings are positioned closer to the nose of the vehicle than to its tail. Consequently, following a small perturbation $\partial \alpha$ that increases the AoA, the wings induce an additional nose-up pitching moment. By contrast, for a narrow range of AoAs, the inflated configurations satisfy the principal condition for static stability,
$$
M_\alpha = \frac{\partial |\bm{M}|}{\partial \alpha} > 0.
\eqno{(8)}
$$
An appropriate mass distribution could therefore reduce the net pitching moment acting on the vehicle and, together with the inflated wings, bring the UUV into a statically stable configuration when required. 
Finally, an analysis of the rolling moment was also conducted to highlight the enhanced manoeuvrability that soft morphing wings confer on UUVs. Conventional vehicles typically rely on eccentric counterweights to induce roll, whereas deformable wings enable direct roll control through differential deformation. The corresponding results are shown in Fig.~\ref{fig:piroll}: soft morphing wings clearly expand the manoeuvring capabilities of UUVs, providing additional means to regulate stability and achieve direct turning. 
Although the responsiveness of this control depends on the time for wing deformation—reported in~\cite{Giordano2024} as under one minute—the proposed actuation strategy is intended for quasi-static trim, stability, and trajectory correction rather than high-bandwidth control. The energetic cost of wing morphing scales with the hydraulic work to move fluid between chambers, roughly the pressure difference times the displaced volume. For the maximum tested inflation (120 mL) and a pressure difference of \SI{50}{\kilo\pascal}, this amounts to about \SI{6}{\joule} per actuation, negligible compared to buoyancy modulation and hydrodynamic drag over a typical mission. By contrast, shifting a counterweight requires mechanical work against gravity and friction; because the proposed strategy does not rely on counterweights, it also relaxes weight constraints on the UUV.
Conventional UUVs often cannot correct for deviations caused by underwater currents—frequently neglected in models—leading to unexpected trajectory drifts~\cite{fan2014dynamics}. Deformable wings, however, enable quasi-continuous correction, enhancing trajectory control at fine spatial scales and allowing precise navigation in confined or complex environments, such as small islands or icebergs.

\begin{figure*}[!tb]
    \centering
    \includegraphics[width=\textwidth]{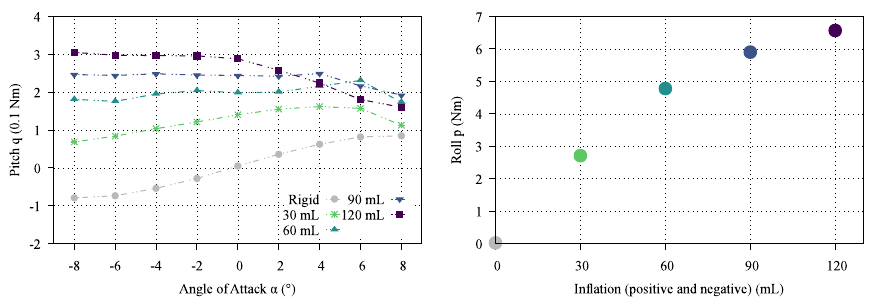}
   \caption{\textit{Left hand side}: simulation of pitching moments $q$ for each inflation tested experimentally at a flow-speed of \SI{0.26}{\metre\per\second}. \textit{Right hand side}: simulation of rolling moments $p$ for all tested inflations at $\alpha = 0\degree$. Each inflation shown in this plot is given under the assumption that the two wings are inflated in opposite directions ($30$ and $-30$ mL, $60$ and $-60$ mL, $90$ and $-90$ mL, and $120$ and $-120$ mL).}
    \label{fig:piroll}
    \vskip -3eX
\end{figure*}
\section{Conclusion}
This study has demonstrated the potential benefits of implementing a camber-morphing wing on UUVs. The soft wing tested in~\cite{Giordano2024} was accurately reproduced in digital form, and its structural deformation was simulated using \textsc{Ansys Mechanical}.
Then, the hydrodynamic performance of a UUV incorporating the soft wing was characterised in 
\begin{figure}[H]
\vskip -2eX
    \centering
            \centering
\includegraphics[width=\columnwidth]{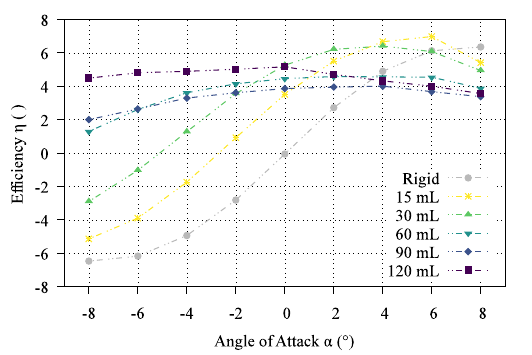}
    \vskip -2eX
    \caption{Hydrodynamic characterization of the selected Underwater Unmanned Vehicle (UUV) implementing the soft wing at all inflations tested in \cite{Giordano2024} and the interpolated \SI{15}{\milli\litre}. Free-flow speed $U_1=\SI[per-mode=symbol]{0.26}{\metre\per\second}$ .}
    \label{fig:efficiency026}
\end{figure}
\noindent \textsc{Ansys Fluent}. The results indicate that the soft morphing wing can enhance the overall efficiency of the vehicle by up to $9.75\%$, thereby significantly extending its operational range and improving manoeuvrability through moment control.
The crucial role of these vehicles in environmental sensing has been emphasised in the literature. For missions in cluttered environments such as the Arctic or the Southern Ocean, where deployment and recovery operations may be impeded by pack ice or icebergs, the operational range becomes a critical factor~\cite{lucas2025giant}. 
Furthermore, this study provides one of the first documented examples of applying soft robotics to gliding vehicles, specifically UUVs that rely on buoyancy-driven forward motion, to enhance their overall performance.
Finally, the present work lays the foundation for future experimental investigations on UUVs implementing soft morphing wings, aimed at collecting comprehensive data on their hydrodynamic, mechanical, and electrical efficiency, and comparing these with conventional underwater gliders.
While the quantitative efficiency gains reported here depend on the specific wing and actuation adopted from \cite{Giordano2024}, the observed trends in efficiency, static stability, and roll authority are general consequences of camber morphing and are expected to extend to other soft wing designs. 
\section{Acknowledgments}
This work was supported in part by the ERC Consolidator
Grant funded by SERI, ProteusDrone grant no. MB22.00066. The authors thank the members of the Laboratory of Sustainability Robotics for constructive discussions along the way, in particular to Dr. Fabian Wiesemüller who helped in scoping and technical input.

\bibliographystyle{IEEEtran}
\bibliography{IEEEabrv, references}

@article{Giordano2024,
  author    = {Andrea Giordano and Liam Achenbach and Daniel Lenggenhager and Fabian Wiesemüller and Roger Vonbank and Claudio Mucignat and André Tristany Farinha and Pham Huy Nguyen and Robert Katzschmann and Sophie F Armanini and Ivan Lunati and Sukho Song and Mirko Kovač},
  title     = {{A Soft Robotic Morphing Wing for Unmanned Underwater Vehicles}},
  journal   = {Advanced Intelligent Systems},
  year      = {2024},
  volume    = {6},
  pages     = {2300702},
  month     = {mar},
  doi       = {10.1002/aisy.202300702},
  publisher = {Wiley-VCH GmbH}
}

@incollection{Katzschmann2016HydraulicSwimming,
    title = {{Hydraulic autonomous soft robotic fish for 3D swimming}},
    year = {2016},
    booktitle = {Springer Tracts in Advanced Robotics},
    author = {Katzschmann, Robert K. and Marchese, Andrew D. and Rus, Daniela},
    pages = {405--420},
    volume = {109},
    publisher = {Springer Verlag},
    doi = {10.1007/978-3-319-23778-7{\_}27},
    issn = {1610742X}
}

@article{wu2024effect,
  title={Effect of stress relaxation on the sealing performance of O-rings in deep-sea hydraulic systems: A numerical investigation},
  author={Wu, Jia-Bin and Li, Li and Wang, Pin-Jian},
  journal={Engineering Science and Technology, an International Journal},
  volume={51},
  pages={101654},
  year={2024},
  publisher={Elsevier}
}

@article{wu2023numerical,
  title={Numerical and experimental study of reciprocating seals in seawater hydraulic variable ballast components for 11,000-m operation},
  author={Wu, Defa and Ma, Yunxiang and Wang, Zhenyao and Min, Hao and Deng, Yipan and Liu, Yinshui},
  journal={Tribology Transactions},
  volume={66},
  number={1},
  pages={92--103},
  year={2023},
  publisher={Taylor \& Francis}
}

@article{javaid2017effect,
  title={Effect of wing form on the hydrodynamic characteristics and dynamic stability of an underwater glider},
  author={Javaid, Muhammad Yasar and Ovinis, Mark and Hashim, Fakhruldin BM and Maimun, Adi et al.},
  journal={International Journal of Naval Architecture and Ocean Engineering},
  volume={9},
  number={4},
  pages={382--389},
  year={2017},
  publisher={Elsevier}
}

@article{lavazza2023strain,
  title={Strain rate, temperature and deformation state effect on Ecoflex 00-50 silicone mechanical behaviour},
  author={Lavazza, Jacopo and Contino, Marco and Marano, Claudia},
  journal={Mechanics of Materials},
  volume={178},
  pages={104560},
  year={2023},
  publisher={Elsevier}
}

@article{singh2017cfd,
  title={CFD approach to modelling, hydrodynamic analysis and motion characteristics of a laboratory underwater glider with experimental results},
  author={Singh, Yogang and Bhattacharyya, SK and Idichandy, VG},
  journal={Journal of Ocean Engineering and Science},
  volume={2},
  number={2},
  pages={90--119},
  year={2017},
  publisher={Elsevier}
}

@inproceedings{ichihashi2008development,
  title={Development of an underwater glider with independently controllable main wings},
  author={Ichihashi, Nobumasa and Ikebuchi, Takuro and Arima, Masakazu},
  booktitle={ISOPE International Ocean and Polar Engineering Conference},
  pages={ISOPE--I},
  year={2008},
  organization={ISOPE}
}

@article{aguilar2020onset,
  title={On the onset of negative lift in a symmetric airfoil at very small angles of attack},
  author={Aguilar-Cabello, Jorge and Gutierrez-Castillo, Paloma and Parras, Luis and del Pino, Carlos and Sanmiguel-Rojas, Enrique},
  journal={Physics of Fluids},
  volume={32},
  number={5},
  year={2020},
  publisher={AIP Publishing}
}

@book{Bianco2003a,
  author    = {Claudio Bianco},
  title     = {Meccanica del volo},
  publisher = {Levrotto \& Bella},
  address   = {Torino},
  year      = {2003},
  isbn      = {978-88-7285-132-2},
  organization      = {Politecnico di Torino}
}

@article{jardin2025naca0012,
  title={NACA0012 airfoil at Reynolds numbers between 50,000 and 140,000—Part 1: Steady freestream},
  author={Jardin, T and Ferrand, V and Gowree, ER},
  journal={International Journal of Heat and Fluid Flow},
  volume={112},
  pages={109655},
  year={2025},
  publisher={Elsevier}
}

@article{suberg2014assessing,
  title={Assessing the potential of autonomous submarine gliders for ecosystem monitoring across multiple trophic levels (plankton to cetaceans) and pollutants in shallow shelf seas},
  author={Suberg, Lavinia and Wynn, Russell B and Van Der Kooij, Jeroen and Fernand, Liam and Fielding, Sophie and Guihen, Damien and Gillespie, Douglas and others},
  journal={Methods in Oceanography},
  volume={10},
  pages={70--89},
  year={2014},
  publisher={Elsevier}
}

@article{fan2014dynamics,
  title={Dynamics of underwater gliders in currents},
  author={Fan, Shuangshuang and Woolsey, Craig A},
  journal={Ocean Engineering},
  volume={84},
  pages={249--258},
  year={2014},
  publisher={Elsevier}
}

@article{lucas2025giant,
  title={Giant iceberg meltwater increases upper-ocean stratification and vertical mixing},
  author={Lucas, Natasha S and Brearley, J Alexander and Hendry, Katharine R and Spira, Theo and Braakmann-Folgmann, Anne and Abrahamsen, E Povl and Meredith, Michael P and Tarling, Geraint A},
  journal={Nature Geoscience},
  pages={1--8},
  year={2025},
  publisher={Nature Publishing Group UK London}
}

@article{hossain2025recent,
  title={Recent advancements in biomimetic application in underwater glider development: Current opportunities and future trends},
  author={Hossain, Gazi Arman and Saimoon, Nayem Zaman and Salehin, Md Nazmus and Ahmed, Mim Mashrur and Masud, Mahadi Hasan},
  journal={Ain Shams Engineering Journal},
  volume={16},
  number={7},
  pages={103410},
  year={2025},
  publisher={Elsevier}
}

\end{document}